\documentclass{article}
\usepackage{authblk}
\title{Comparison of Autoencoders for tokenization of ASL datasets}
\author[1]{Vouk Praun-Petrovic}
\author[2]{Aadhvika Koundinya}
\author[3]{Lavanya Prahallad}
\affil[1]{Harker Upper School, San Jose,  California}
\affil[2]{Irvington High School, Fremont, California.}
\affil[3]{Research Spark Hub, Dublin, California.}

\usepackage{graphicx}
\usepackage{booktabs}
\usepackage[backend=bibtex,style=numeric]{biblatex}
\bibliography{references}
\begin{document}

\maketitle
\section{Introduction}
Generative AI powered by large language models (LLMs) has revolutionized a wide array of applications, ranging from chatbots and virtual assistants to the generation of sound, audio, and video. With advancements in multimodal AI, LLMs are increasingly capable of handling tasks involving text, audio, images, and video \cite{bengesi2023advancements}. Among these, applications such as sign language detection and generation have gained prominence, leveraging LLMs' ability to process and integrate data across different modalities. These multimodal models employ encoder-decoder mechanisms as essential components to preprocess input data and post-process outputs for various tasks. These mechanisms ensure that raw input data—whether visual, audio, or textual—is converted into a tokenized form that the LLM can process, and then decoded into meaningful output.

The quality and precision of encoder-decoder mechanisms play a pivotal role in the success of multimodal applications. An effective encoder-decoder framework not only preserves the semantic and structural information of the input during encoding but also ensures high-fidelity reconstruction during decoding. In multimodal applications, such as sign language generation, inaccuracies in encoding can lead to the loss of critical contextual or spatial information, while errors in decoding can result in outputs that fail to convey the intended message. Thus, designing and evaluating robust encoding and decoding systems is essential to enable multimodal LLMs to perform reliably across diverse tasks.

This work focuses specifically on developing encoder-decoder models for an American Sign Language (ASL) image dataset—a challenging yet essential task for enabling effective communication between individuals with hearing impairments and digital systems. The ASL dataset presents a rich set of hand sign images representing 29 distinct classes, including letters, space, delete, and nothing. By addressing this dataset, we aim to create models that can efficiently tokenize image data for LLMs and reconstruct images with high accuracy. Through this work, we contribute to the advancement of encoder-decoder architectures tailored for visual modalities, a critical component in multimodal AI systems \cite{wikipedia_gan}.

To evaluate the effectiveness of different approaches, we compare three encoder-decoder techniques: a simple feedforward network, a convolutional network, and a diffusion network. Each technique represents a distinct methodology for encoding and decoding image data \cite{wikipedia_transformer}. The feedforward network provides a baseline with fully connected layers operating on flattened image inputs. The convolutional network leverages spatial hierarchies in image data through convolutional and transposed convolutional layers, making it more effective for image reconstruction. The diffusion network introduces probabilistic modeling and iterative noise prediction into the encoding and decoding process, adding robustness to noise and improving generative capabilities \cite{wei2023diffusionmae}.

Our study offers a comprehensive analysis of these techniques through both objective and subjective evaluations. Objective metrics such as reconstruction error and accuracy provide quantifiable insights into the performance of each model, while subjective evaluations assess the perceptual quality of the reconstructed images. By comparing these approaches, this work highlights the strengths and weaknesses of each technique and provides valuable insights for developing robust encoder-decoder systems for multimodal AI applications, particularly in the context of sign language recognition and generation.

\section{Data}
\subsection{Data Description}
The dataset used in this study is the ASL Alphabet Dataset sourced from Kaggle, comprising 87,000 images across 29 classes of hand signs. These classes represent the letters A-Z, along with signs for “space” and “delete,” as well as a "nothing" class of blank images. Each class contains 3,000 images, ensuring diverse representations with variations in hand sizes, lighting conditions, and backgrounds. This diversity makes the dataset well-suited for training robust models capable of recognizing hand signs in varying real-world conditions.

To facilitate training and evaluation, the dataset was split into an 80/20 ratio, yielding 69,600 images for training and 17,400 for validation. Additionally, a test set containing a single image for each class, excluding "delete" (28 images total), was created to evaluate the model's performance on unseen data.
\subsection{Data Pre-Processing}
Prior to feeding the data into the autoencoders, several preprocessing steps were performed to ensure consistency and suitability for training. The images were first resized to a fixed resolution of 200×200 pixels to match the input requirements of the models. Each image was then converted into a tensor, a multidimensional data structure used by PyTorch, and normalized to scale pixel values to the range [0, 1]. This normalization step ensures uniformity across the dataset, preventing issues arising from varying pixel intensity scales. These normalized images were then separated into training and validation datasets, with an 80/20 split. By applying these preprocessing techniques, the dataset was prepared to efficiently train the autoencoders while preserving critical features necessary for the accurate representation and reconstruction of hand signs \cite{preechakul2022diffusionae}.

\section{Models}
\subsection{Feedforward Autoencoder}
The Feedforward Autoencoder is a fully connected neural network designed to compress high-dimensional input data into a lower-dimensional latent representation and reconstruct it back to its original form. It treats the input as a flattened vector, making it suitable for simpler datasets or tasks that do not require spatial feature extraction. Its straightforward structure makes it a strong baseline for understanding autoencoding techniques.

\subsection{Details}
The Feedforward Autoencoder is a fully connected neural network designed to compress high-dimensional input data into a lower-dimensional latent representation and reconstruct it back to its original form. It treats the input as a flattened vector, making it suitable for simpler datasets or tasks that do not require spatial feature extraction. Its straightforward structure makes it an effective baseline for understanding autoencoding techniques.

\begin{figure}[htbp]
    \centering
    \includegraphics[width=0.5\textwidth]{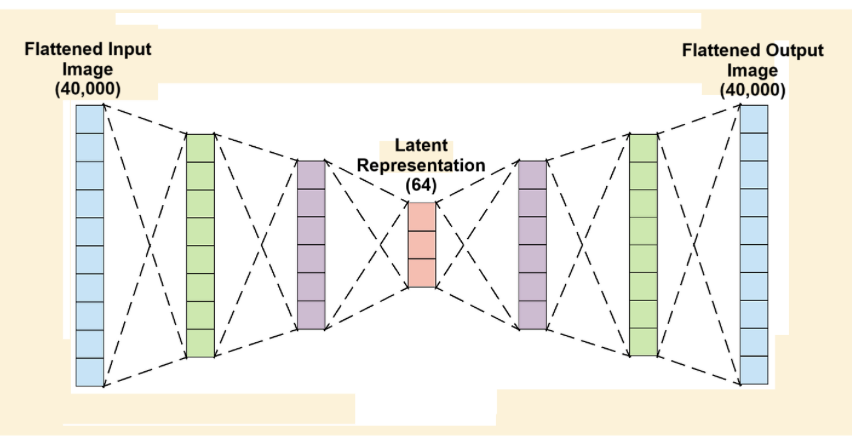}
    \caption{Visual Depiction of Feedforward Autoencoder architecture}
    \label{fig1}
\end{figure}

\subsection{Convolutional Autoencoder}
The Convolutional Autoencoder utilizes convolutional layers to encode and decode image data while preserving the spatial structure of the input. It is particularly well-suited for image-related tasks, as convolutional operations hierarchically extract spatial features such as edges, textures, and complex patterns.

\subsection{Details}
The encoder consists of three convolutional layers, each using 4×4 filters with a stride of 2 and padding of 1. These layers progressively reduce the spatial dimensions of the input image from 200×200 to 100×100, 50×50, and finally 25×25, while increasing the number of feature maps from 32 to 64 and 128. After this, the encoder flattens the 25×25×128 tensor and applies a linear layer to reduce its size to 64 dimensions. The decoder reverses this process using another linear layer followed by transposed convolutional layers, restoring the latent representation to the original image size. Each convolutional and transposed convolutional layer employs ReLU activations, except for the final output layer, which uses a sigmoid activation function. This network contains approximately 16 million parameters, making it significantly more efficient than the feedforward autoencoder due to parameter sharing in the convolutional layers. Its ability to capture hierarchical spatial features makes it highly effective for image reconstruction tasks.

\begin{figure}[htbp]
    \centering
    \includegraphics[width=0.5\textwidth]{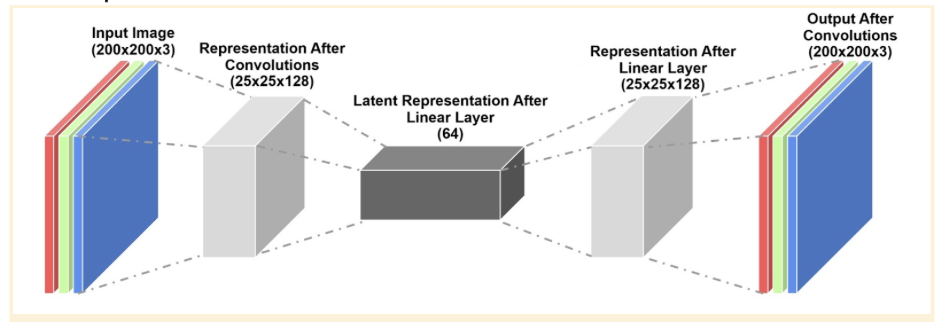}
    \caption{Visual Depiction of Convolutional Autoencoder architecture}
    \label{fig2}
\end{figure}

\subsection{Diffusion Autoencoder}
The Diffusion Autoencoder extends the convolutional autoencoder by incorporating a diffusion process, adding probabilistic noise, and performing denoising during the encoding phase. This architecture combines principles of variational autoencoders with iterative noise prediction, making it robust for tasks involving noise or generative data modeling \cite{bounoua2023multimodal, preechakul2022diffusionae}.

\subsection{Details}
The encoder retains the structure of the convolutional autoencoder, with three convolutional layers reducing the spatial dimensions and producing a latent representation characterized by mean and variance through fully connected layers. The diffusion model introduces a noise schedule over 100 timesteps, perturbing the latent variables using a beta schedule. A denoising model, consisting of a fully connected network with five primary layers, multiple dynamically added layers, and final output layers, predicts and removes the noise during reverse diffusion \cite{wikipedia_stablediffusion}. The decoder, mirroring the convolutional structure, reconstructs the input image from the denoised latent space. With approximately 16 million parameters, the Diffusion Autoencoder demonstrates its strength in generating high-quality reconstructions under noisy conditions and is particularly valuable for robust generative tasks and probabilistic modeling  \cite{sohldickstein2015deep}.

\begin{figure}[htbp]
    \centering
    \includegraphics[width=0.5\textwidth]{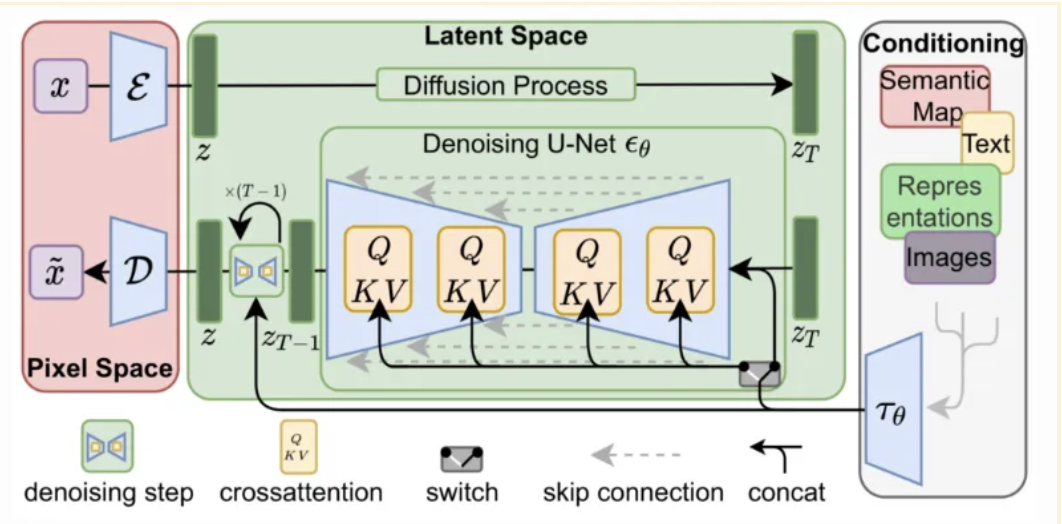}
    \caption{Visual Depiction of noise schedule and de-noising model}
    \label{fig3}
\end{figure}

\section{Model Building}
The training process begins by initializing the chosen model (Diffusion, Convolutional, or Feedforward Autoencoder) and preparing it for optimization using the Adam optimizer. We utilized 500 images per class, which were divided into training and validation datasets using an 80/20 split. A mean squared error (MSE) loss function is employed to measure reconstruction accuracy by comparing the model's output to the original input images. The model is trained over multiple epochs, iterating through batches of images in the training dataset. During each iteration, the input images are passed through the model, and the loss is computed. For the Diffusion Autoencoder, additional intermediate outputs, such as latent variables and noise, are generated during the forward pass, while simpler feedforward and convolutional models directly reconstruct the input. The loss is backpropagated through the network to update the model's parameters via gradient descent. After training, the model's architecture and weights are saved for future evaluation or deployment, ensuring reproducibility and flexibility for further experiments.

\section{Evaluation}
\subsection{Evaluation Methodology}
The evaluation of the encoder-decoder models involves both subjective and objective approaches, ensuring a comprehensive assessment of their performance. The objective evaluation metric is the mean squared error (MSE) of each model across the validation dataset, which consists of 2,900 images randomly selected from the dataset using a fixed random seed. For the subjective evaluation, a method is designed to gauge human perception of the quality of reconstructed images. In this process, five human subjects are presented with reconstructed images generated by the models and are asked to rate them based on their visual fidelity and resemblance to the original input images. The participants are provided with a straightforward evaluation scale, ranging from 1 to 5, where 1 represents "Bad," 2 represents "Poor," 3 represents "Fair," 4 represents "Good," and 5 represents "Excellent." This approach captures qualitative feedback that reflects how effectively the models reconstruct images from a human perspective, which is crucial for tasks such as sign language recognition, where visual clarity directly impacts usability.

To ensure a fair and unbiased subjective evaluation, the images are presented to participants without revealing the model used to generate them. Participants are given clear instructions and examples of the rating scale to standardize their evaluations. This subjective methodology complements objective metrics, such as reconstruction loss, by providing insights into the perceptual quality of the outputs \cite{liang2023multimodal}.

\section{Results}

\begin{figure}[htbp]
    \centering
    \includegraphics[width=0.5\textwidth]{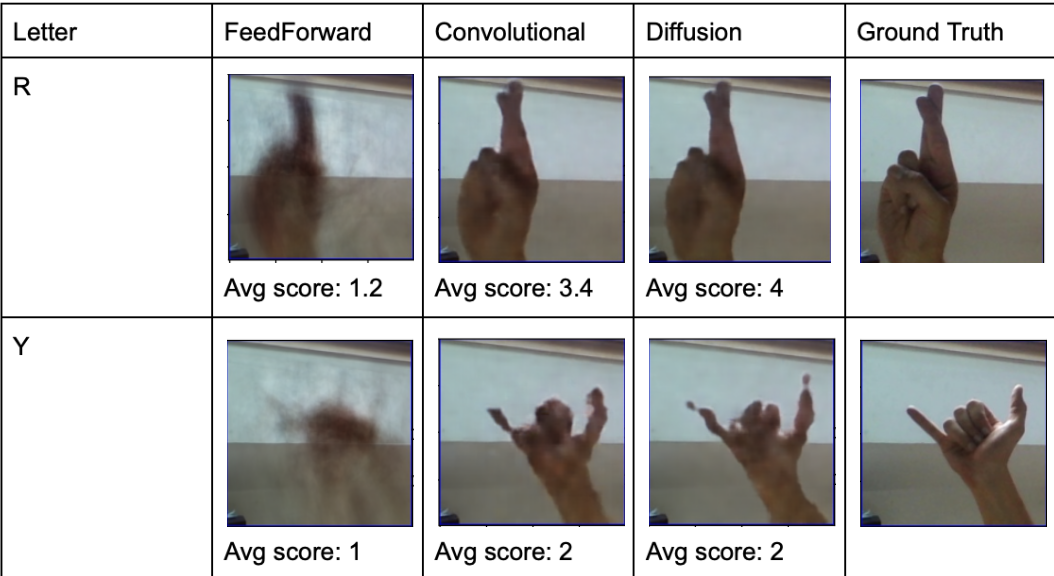}
    \caption{Sample Reconstructions of Each Architecture and their Average Mean Opinion Scores (MOS)}
    \label{Table1}
\end{figure}

As shown above, the feedforward network struggles to recreate high-quality images, as the hands in both images appear unclear, and the autoencoder introduces blurriness throughout the image. Consequently, its average Mean Opinion Score (MOS) from human reviewers for both classes was lower. In contrast, the convolutional and diffusion networks performed significantly better in reconstructing the letter R. A marginal improvement in the clarity of the fingers is evident in the diffusion network’s reconstruction of the letter R, which explains its higher average MOS. However, both networks faced greater difficulty with the hand sign for Y. The varied lighting on the sign, along with the fingers being more spread out, likely contributed to the lower quality of these reconstructions and the corresponding drop in their average MOS. 

\subsubsection{Results of Subjective Evaluation}
\begin{table}[htbp]
    \centering
    \begin{tabular}{|l|c|}
        \hline
        \textbf{Method} & \textbf{Average Mean Opinion Score} \\ \hline
        FeedForward     & 1.28 \\ \hline
        Convolution     & 2.79 \\ \hline
        Diffusion       & 3.11 \\ \hline
    \end{tabular}
    \caption{MOS for Reconstructed Images Across Different Encoder-Decoder Models}
    \label{Table}
\end{table}
The results of the subjective evaluation indicate that the diffusion network outperformed the other two models, achieving an average Mean Opinion Score (MOS) of 3.11. The Convolutional Network followed with a score of 2.79, while the Feedforward Network scored the lowest at 1.28.
\subsubsection{Objective Results}

\begin{table}[htbp]
    \centering
    \begin{tabular}{|l|c|}
        \hline
        \textbf{Method} & \textbf{MSE (Mean Squared Error)} \\ \hline
        FeedForward     & 0.00506 \\ \hline
        Convolution     & 0.00144 \\ \hline
        Diffusion       & 0.00141 \\ \hline
    \end{tabular}
    \caption{MSE for Reconstructed Images Across Different Encoder-Decoder Models on Validation Dataset}
    \label{Table}
\end{table}

The objective results display a trend similar to that observed for the Mean Opinion Score (MOS). The diffusion network slightly outperformed the convolutional network, achieving scores of 0.00141 and 0.00144, respectively, while the feedforward network lagged behind by a significant margin. This indicates that the performance of the models, as evaluated by the Mean Squared Error (MSE) criterion, aligns closely with human perceptions of quality.

These results highlight the strengths and weaknesses of each model's architecture in reconstructing images. The higher score of the diffusion network suggests that its iterative noise modeling and probabilistic framework enable more accurate and perceptually pleasing reconstructions, even under challenging conditions such as noisy or diverse datasets \cite{liu2023generalized}. This makes it particularly suitable for tasks where robustness and high-quality visual output are critical, such as sign language interpretation.

In contrast, the convolutional network's performance reflects its ability to effectively capture spatial features through convolutional layers, enabling decent reconstructions that maintain the structural integrity of the input images. However, it falls short of the diffusion network, likely due to the absence of probabilistic noise correction mechanisms. The feedforward network's lower score demonstrates its limitations in handling image data, as its fully connected structure does not exploit the spatial correlations inherent in images. This results in less visually accurate reconstructions compared to the other two models.

Overall, these findings emphasize the superiority of the diffusion network for high-fidelity image reconstructions, with the convolutional network serving as a viable intermediate option and the feedforward network acting as a baseline.

\section{Summary and Conclusions}
This study explored the development and evaluation of encoder-decoder architectures for the American Sign Language (ASL) image dataset, comparing three distinct approaches: Feedforward Autoencoder, Convolutional Autoencoder, and Diffusion Autoencoder. The Diffusion Autoencoder demonstrated superior performance, achieving the highest subjective evaluation score by leveraging its iterative noise modeling and probabilistic framework to produce high-quality image reconstructions. The Convolutional Autoencoder performed moderately well, effectively capturing spatial hierarchies in image data but lacking the robustness provided by the diffusion process. The Feedforward Autoencoder served as a baseline, with its fully connected structure proving less effective for handling image data due to its inability to exploit spatial features, which further explains its lower performance.

These results underscore the importance of incorporating spatial and probabilistic modeling for applications involving structured data, such as images. The findings highlight the potential of Diffusion Autoencoders for robust multimodal AI applications, particularly for tasks such as sign language recognition and generation, where accurate and high-quality reconstructions are critical.

\printbibliography

\end{document}